\documentclass[letterpaper, 10 pt, conference]{ieeeconf} 

\IEEEoverridecommandlockouts 
\usepackage{graphicx}
\usepackage{amsmath}
\usepackage{arydshln}
\usepackage{hhline}
\usepackage{eucal}
\usepackage{cite}
\usepackage{amssymb}
\usepackage{enumerate}
\usepackage[caption=false,font=footnotesize]{subfig}
\usepackage{bm}
\usepackage{color}
\usepackage{multirow}
\usepackage{algorithm} 
\usepackage{algpseudocode} 
\usepackage{graphicx}
\usepackage{svg}
\usepackage{booktabs}
\usepackage{lipsum}
\usepackage{mathtools}

\usepackage{atbegshi,picture}
\usepackage{lipsum}
\usepackage{textcomp}
\newcommand{\gray}[1]{{\color{gray}#1 }}
\AtBeginShipout{\AtBeginShipoutUpperLeft{%
  \put(\dimexpr\paperwidth/2\relax,-1.25cm){\makebox[0pt][c]{
  \begin{tabular}{c}%
    \gray{This paper has been accepted for publication at the} \\ \gray{IEEE International Conference on Robotics and Automation (ICRA), Yokohama, 2024. \textcircled{c}IEEE} %
  \end{tabular}}%
  }}%
}

\makeatletter
\renewcommand{\ALG@beginalgorithmic}{\small}
\makeatother

\providecommand{\norm}[1]{\lVert#1\rVert}

\makeatletter
\renewcommand{\env@matrix}[1][c]{%
  \hskip-\arraycolsep
  \let\@ifnextchar\new@ifnextchar
  \array{*\c@MaxMatrixCols #1}%
}
\makeatother
\newcommand{\argmin}{\arg\!\min} 

\DeclareSymbolFontAlphabet{\mathcal}   {symbols}

\title{\LARGE \bf
Online Multi-Contact Feedback Model Predictive Control \\ for Interactive Robotic Tasks
}

\author{Seo Wook Han$^{1}$, Maged Iskandar$^{2}$, Jinoh Lee$^{2,3,\dagger}$ and Min Jun Kim$^{1,\dagger}$
\thanks{
This work was supported by ITECH R$\&$D program of MOTIE/KEIT under Project No. 20014398 and No. 20014485), 
and by the National Research Foundation of Korea (NRF) grant funded by the Korea government (MSIT) (Project No. 2021R1C1C1005232 and No. 2021R1A4A3032834).}
\thanks{
\textsuperscript{1}The authors are with Intelligent Robotic Systems Laboratory, Korea Advanced Institute of Science and Technology (KAIST), Daejeon, Republic of Korea. {E-mail: {\tt\small tjdnr7117, minjun.kim@kaist.ac.kr}}}
\thanks{
\textsuperscript{2}The authors are with Institute of Robotics and Mechatronics, German Aerospace Center (DLR), Wessling, Germany. {E-mail: {\tt\small {first name.last name}@dlr.de}}}
\thanks{
\textsuperscript{3}The author is also with the Department of Mechanical Engineering, KAIST, Daejeon, Republic of Korea.
}
\thanks{
\textsuperscript{$\dagger$}Co-corresponding authors.
}
}

\begin{document}

\maketitle
\thispagestyle{empty}
\pagestyle{empty}

\begin{abstract}
In this paper, we propose a model predictive control (MPC) that accomplishes interactive robotic tasks, in which multiple contacts may occur at unknown locations. To address such scenarios, we made an explicit contact feedback loop in the MPC framework. An algorithm called Multi-Contact Particle Filter with Exploration Particle (MCP-EP) is employed to establish real-time feedback of multi-contact information. Then the interaction locations and forces are accommodated in the MPC framework via a spring contact model. Moreover, we achieved real-time control for a 7 degrees of freedom robot without any simplifying assumptions by employing a Differential-Dynamic-Programming algorithm. We achieved $6.8\mathrm{kHz}$, $1.9\mathrm{kHz}$, and $1.8\mathrm{kHz}$ update rates of the MPC for 0, 1, and 2 contacts, respectively. This allows the robot to handle unexpected contacts in real time. Real-world experiments show the effectiveness of the proposed method in various scenarios.
\end{abstract} 

\section{Introduction}
\label{sec:introduction}

Physical interaction capabilities are crucial in interactive robotic tasks.
Especially for collaborative robots, which are designed to share a workspace with humans, there is an increased need for effective handling of physical interactions.
For example, table polishing requires the robot to accomplish hybrid motion-force control.
Moreover, human-robot collaboration scenarios may involve multiple unknown contacts, which need to be safely managed in such a way that the interaction force is maintained below a dangerous level.

There have been a number of attempts to enable robots to perform such interactive tasks.
For example, a hybrid control method that simultaneously realizes both motion and force control has been proposed in \cite{iskandar2023hybrid}.
Additionally, to ensure safe interaction, compliant behavior of robots through impedance or admittance control has been implemented in \cite{ott2010unified, kim2019model, 9812268 ,9643422, 7759143, keemink2018admittance, 8695069}.

\begin{figure}[tb!]
\centering
	{\includegraphics[width=0.475\textwidth, trim = {5px 5px 0 0px}, clip]{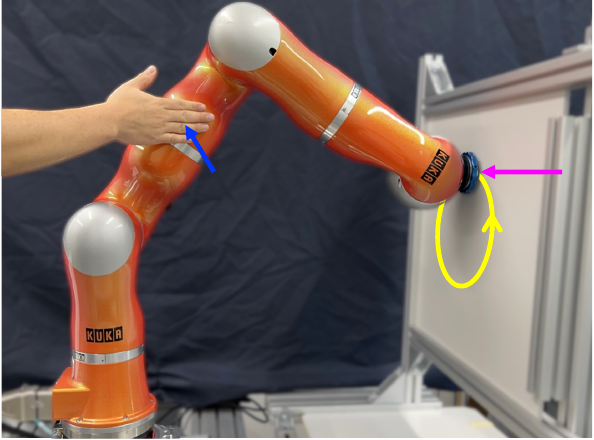}} 
	\caption{
        The simultaneous accomplishment of hybrid motion-force control at the end-effector (yellow and magenta colored arrows) and safe handling of unexpected contact caused by human interaction (blue colored arrow).
        }
	\label{fig:introduction_problem}
	\vspace{-6mm}
\end{figure}

With recent advancements in optimization techniques,
there have been attempts to address interactive robotic tasks using model predictive control (MPC).
By utilizing the MPC, such interaction tasks can be accomplished through cost function designs \cite{tassa2012synthesis}, while imposing constraints on the state and input variables \cite{tassa2014control, mastalli2022feasibility, aoyama2021constrained}.

To use the MPC framework for interaction tasks, a contact model should be defined to predict the contact-involved robot motion in the prediction horizon.
While approaches based on kinematic constraints show promising results \cite{kleff2022introducing, mastalli2020crocoddyl}, other types of contact models are also being explored.
Particularly, a spring contact model, in which force feedback can be incorporated, is often employed \cite{kazim2018combined, wahrburg2016mpc, bednarczyk2020model, gold2020model, gold2021model, gold2022model, matschek2017force, matschek2020direct, matschek2023safe}; e.g., interaction tasks are realized via admittance or impedance behavior of the robot with constraints on the contact force and robot state \cite{wahrburg2016mpc, kazim2018combined, bednarczyk2020model}, and studies in \cite{matschek2017force, matschek2020direct, matschek2023safe,  gold2021model, gold2022model} have reported successful results for hybrid motion-force control.

Nevertheless, the aforementioned methods are valid based on a common assumption that the contact occurs at a single and known location, typically at the end-effector where a force/torque (F/T) sensor is installed.
This assumption seems reasonable in a well-structured workspace because all contacts can be planned in advance.
However, in an unstructured workspace, such assumptions clearly have limitations.
For example, in scenarios that involve physical interaction with humans or with unknown obstacles, unexpected contact may occur at multiple locations on the robot.

To tackle the problems of unknown contacts, algorithms that simultaneously estimate the contact location and force have been proposed \cite{iskandar2021collision, manuelli2016localizing, 10161173}.  
Some studies have presented control methods that can safely handle a single contact using those algorithms \cite{pang2022easing,zube2016model}. 
For example, authors in \cite{pang2022easing} use a kinematic constraint-based contact model, to generate a path that maintains the contact force below the pre-specified value, and in \cite{zube2016model}, a collision avoidance path is generated along the force direction.
Moreover, the safe handling of multiple contacts has been demonstrated using tactile sensors \cite{8695719, 8932392}.
For example, authors in \cite{jain2013reaching, killpack2016model} utilize tactile sensor feedback to minimize resulting multiple contact forces in contact-rich environments. 
However, it is not straightforward how to handle unknown multiple contacts, occurring outside the tactile sensor, although such capability is crucial in accomplishing complex interaction tasks.

Accordingly, this paper proposes a {\it contact-feedback MPC framework} that addresses complex interaction tasks in which {\it multiple contacts} may occur {\it at arbitrary locations} (see, e.g., Fig. \ref{fig:introduction_problem}).
To the best of our knowledge, this is the first attempt that embeds multiple unknown contacts in the MPC framework.
The multi-contact feedback is provided through our previous work called Multi-Contact Particle Filter with Exploration Particles (MCP-EP) \cite{10161173}.
This method is capable of simultaneously estimating multiple contact locations and forces, regardless of the contact links.
The outline of this algorithm is described in Section~\ref{subsec:mcp_ep}.

Compared to the prior work, the proposed multi-contact feedback MPC framework offers several advantages:
(i) By making an explicit contact feedback loop using MCP-EP, force control can be accomplished even in scenarios with multiple contacts regardless of contact locations\footnote{To avoid misunderstanding, it should be mentioned that the MCP-EP algorithm can estimate up to one contact per link.}.
(ii) By employing the Differential Dynamic Programming (DDP) algorithm, the proposed MPC operates in real-time for a 7 degrees of freedom (DOF) manipulator without any simplifying assumptions on the robot model, which are common in literature \cite{wahrburg2016mpc, bednarczyk2020model, gold2020model, gold2021model, gold2022model, zube2016model,jain2013reaching, killpack2016model, pang2022easing}. 
For example, some of the time derivatives are assumed to be constant over the prediction horizon in \cite{gold2021model,gold2022model}, and \cite{pang2022easing} employs a quasi-static robot assumption.
Additionally, a linearized robot model is employed in \cite{bednarczyk2020model,killpack2016model}, and
the robot joints are assumed to be perfectly velocity controlled in \cite{zube2016model, wahrburg2016mpc},
or position controlled in \cite{jain2013reaching}.
In fact, we were able to achieve $6.8\mathrm{kHz}$, $1.9\mathrm{kHz}$, and $1.8\mathrm{kHz}$ when there are 0, 1, and 2 contacts, respectively.
This feature enables the handling of multiple contacts in real-time, as illustrated in Fig. \ref{fig:introduction_problem}.

This paper is organized as follows.
Section~\ref{sec:contact_involved_robot_dynamics} presents the robot system and contact model.
The proposed contact-feedback MPC framework is presented in Section~\ref{sec:optimal_control_problem}.
In Section~\ref{sec:experiments}, the proposed MPC framework is validated through real-world experiments with a DLR-KUKA LWR IV+.

\section{System and Contact Modeling}
\label{sec:contact_involved_robot_dynamics}
To predict the robot's motion over the prediction horizon of the MPC, the system model is given as 
\begin{align}
\label{eq:robot_state_dyn}
&\dot{\boldsymbol{x}}(t) = \boldsymbol{f}(\boldsymbol{x}(t),\boldsymbol{u}(t), \boldsymbol{\lambda}(t)), & &\boldsymbol{x}(0) = \tilde{\boldsymbol{x}},
\end{align}
where $\boldsymbol{x}(t)\in\mathbb{R}^{n_x}$ is the state of the system,  $\boldsymbol{f}$ is the contact-involved robot dynamics,  $\boldsymbol{u}(t)\in\mathbb{R}^{n_u}$ is the control input, and $\tilde{\bullet}$ represents the initial value of $\bullet$, while 
$\boldsymbol{\lambda}(t)$ represents $k$ contact forces, i.e., $\boldsymbol{\lambda}(t) = \{\boldsymbol{\lambda}_{1}(t), \ldots , \boldsymbol{\lambda}_{k}(t)\} \in \mathbb{R}^{3 \times k}$,
of which the $i^{\text{th}}$ element can be expressed as follows:
\begin{align}
\label{eq:contact_state_dyn}
\boldsymbol{\lambda}_{i}(t) = \boldsymbol{g}(\boldsymbol{x}(t);\boldsymbol{\theta}_{i}), \; \boldsymbol{\theta}_{i} = \boldsymbol{h}(\boldsymbol{\tilde{r}}_{c,i},\boldsymbol{\tilde{\lambda}}_{i}),  \; \text{\scriptsize $\forall i = \{1,2, \ldots , k\}$},
\end{align}
where the subscript $i$ denotes the index of $k$ contacts, 
$\boldsymbol{g}$  is the spring contact model,
$\boldsymbol{\theta}_{i}$ is the parameter required for the spring contact model, 
$\boldsymbol{r}_{c,i}(t)\in\mathbb{R}^{3}$ is the position vector of the $i^{\text{th}}$ contact point,
and $\boldsymbol{h}$ represents the procedure of computing $\boldsymbol{\theta}_i$ based on the $i^{\text{th}}$ contact feedback ($\boldsymbol{\tilde{r}}_{c,i}$ and $\boldsymbol{\tilde{\lambda}}_{i}$).
For clarity, $\boldsymbol{r}_{c,i}(t)$ and $\boldsymbol{\lambda}_{i}(t)$ are represented with respect to the world frame, if not specified otherwise.

We point out that the contact model (\ref{eq:contact_state_dyn}) does not depend on the control input.
This type of contact model allows the MPC framework to incorporate contact force feedback by avoiding an algebraic loop between $\boldsymbol{\lambda}_{i}(t)$ and $\boldsymbol{u}(t)$ \cite{kleff2022introducing}. 
Details of the contact model are described in Section~\ref{subsec:spring_dyn}.
For the sake of brevity, the time argument of quantities is omitted in the following.

\subsection{Contact-involved robot dynamics}
\label{subsec:robot_dyn}
Consider the following articulated multi-body dynamics: 
\begin{align}
\label{eq:robot_dyn}
\boldsymbol{M}(\boldsymbol{q})\ddot{\boldsymbol{q}} + \boldsymbol{C}(\boldsymbol{q},\dot{\boldsymbol{q}})\dot{\boldsymbol{q}} + \boldsymbol{g}(\boldsymbol{q}) = \boldsymbol{\tau}_c + \boldsymbol{\tau}_{ext},
\end{align}
where $\boldsymbol{q}\in\mathbb{R}^{n}$ is the joint variable, $\boldsymbol{M}(\boldsymbol{q})$ is the inertia matrix, $\boldsymbol{C}(\boldsymbol{q},\dot{\boldsymbol{q}})$ is the Coriolis/centrifugal matrix, $\boldsymbol{g}(\boldsymbol{q})$ is the gravity vector, $\boldsymbol{\tau}_c$ is the commanded joint torque, and $\boldsymbol{\tau}_{ext}$ is the external joint torque caused by the contact forces. 

Since we consider scenarios with $k$ contacts, the external joint torque $\boldsymbol{\tau}_{ext}$ is given as
\begin{align}
\label{eq:external_torque}
\boldsymbol{\tau}_{ext} = \sum_{i=1}^{k} \boldsymbol{J}_{i}^{T}(\boldsymbol{q}, \boldsymbol{r}_{c,i}) \boldsymbol{\lambda}_{i},
\end{align}
where $\boldsymbol{J}_{i}(\boldsymbol{q},\boldsymbol{r}_{c,i}) \in \mathbb{R}^{3 \times n}$ is the associated positional Jacobian matrix at the $i^{\text{th}}$ contact point.
Using (\ref{eq:robot_dyn}) and (\ref{eq:external_torque}), we can express the contact-involved robot dynamics of (\ref{eq:robot_state_dyn}) by setting $\boldsymbol{x}=(\boldsymbol{q,\dot{q}})$ and $\boldsymbol{u}=\boldsymbol{\tau}_c$. 

\subsection{Multi-contact feedback}
\label{subsec:mcp_ep}
The initial states of the multiple contacts ($\boldsymbol{\tilde{r}}_{c,i}$ and $\boldsymbol{\tilde{\lambda}}_{i}$) are required for the contact model in (\ref{eq:contact_state_dyn}).
Accurate real-time estimation of these variables is essential to handle unexpected contacts.
In our previous work \cite{10161173}, we proposed a contact estimation algorithm, MCP-EP.
This algorithm can estimate the multiple contacts in a robot manipulator using proprioceptive sensors, e.g., an F/T sensor mounted at the base and joint torque sensors (JTSs).
MCP-EP is suitable for providing contact feedback for the following reasons:
(i) The run-time of MCP-EP is $2200\mathrm{Hz}$ for single-contact and $600\mathrm{Hz}$ for dual-contact; 
(ii) The average root mean squared error (RMSE) of the contact location is $0.16\mathrm{cm}$, and that of the contact force is $0.01\mathrm{N}$, for single-contact.
For dual-contact, these values are $1.08\mathrm{cm}$ and $2.00\mathrm{N}$, respectively.
These numbers are based on a quantitative evaluation of the algorithm in a simulation.
Given these capabilities, in the remainder of this paper, we consider $\boldsymbol{\tilde{r}}_{c,i}$ and $\boldsymbol{\tilde{\lambda}}_{i}$ to be available.

It is worth noting that the proposed MPC does not strictly rely on MCP-EP, while one can employ any method as long as multi-contact information is provided in real time.
The remaining step is to define the contact model and $\boldsymbol{\theta}_i$ in (\ref{eq:contact_state_dyn}) using the contact feedback.

\begin{figure}[tb!]
\centering
        \subfloat[] 
	{\includegraphics[width=0.23\textwidth, trim = {0 2px 0 2px}, clip]{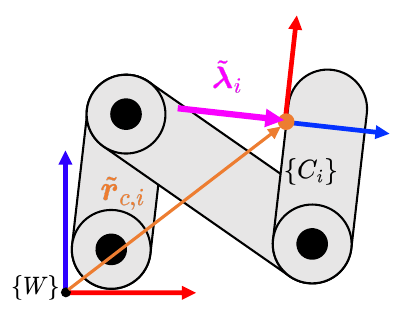}} 
	\subfloat[]
	{\includegraphics[width=0.23\textwidth, trim = {0 26px 0 2px}, clip]{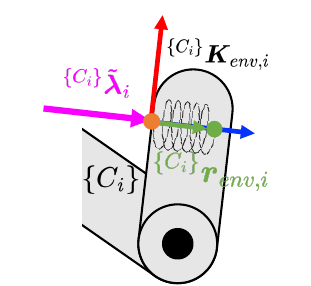}}\\
        \subfloat[]
	{\includegraphics[width=0.23\textwidth, trim = {0 2px 0 7px}, clip]{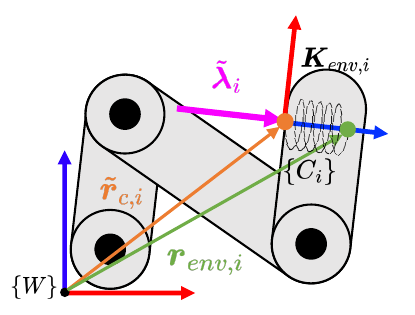}}
	\subfloat[]
	{\includegraphics[width=0.23\textwidth, trim = {0 2px 4px 2px}, clip]{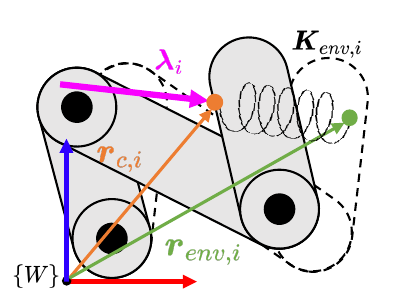}} 
	\caption{ 
    Red and blue colored arrows indicate the $x$ and $z$ axes, respectively, of the frame. 
    (a) The $i^{\text{th}}$ contact frame $\{ C_i\}$ is obtained using the contact feedback $\tilde{\boldsymbol{r}}_{c,i}$ and $\tilde{\boldsymbol{\lambda}}_{i}$.
    (b) $^{\{ C_i \}}\boldsymbol{r}_{env,i}$ is computed using (\ref{eq:env_location_contact_frame}) within the contact frame.
    (c) $\boldsymbol{r}_{env,i}$ and $\boldsymbol{K}_{env,i}$ are calculated by transforming coordinates from $\{ C_i\}$ to $\{ W\}$.
    (d) $\boldsymbol{r}_{c,i}$ is fixed on the robot surface (see the orange dot representing $\boldsymbol{r}_{c,i}$).
    The predictive contact force is derived from $\boldsymbol{\lambda}_{i} = \boldsymbol{g}(\boldsymbol{x};\boldsymbol{\theta}_i)$.
	}
	\label{fig:spring_contact_feedback}
	\vspace{-6mm}
\end{figure}

\subsection{Spring contact model}
\label{subsec:spring_dyn}
We use the following simple spring model, in which the $i^{\text{th}}$ contact force is determined by the stiffness and deformation of the environment:
\begin{align}
\label{eq:spring_dyn_1}
\boldsymbol{\lambda}_{i} = \boldsymbol{K}_{env,i}\Delta{\boldsymbol{r}}_{i},
\end{align}
where $\boldsymbol{K}_{env,i} \in \mathbb{R}^{3\times3}$ and $\Delta{\boldsymbol{r}}_{i} = \boldsymbol{r}_{env,i}-\boldsymbol{r}_{c,i} \in \mathbb{R}^3$ are the $i^{\text{th}}$ environment stiffness matrix and deformation, respectively, and $\boldsymbol{r}_{env,i}$ is the $i^{\text{th}}$ environment location. 
The variables associated with the environment are defined as $\boldsymbol{\theta}_{i} = \{ \boldsymbol{K}_{env,i}, \boldsymbol{r}_{env,i}\}$. 

The procedure of computing $\boldsymbol{\theta}_{i}$ is illustrated in Fig. \ref{fig:spring_contact_feedback}.
As shown in Fig. \ref{fig:spring_contact_feedback}(a), the $i^{\text{th}}$ contact frame $\{ C_i \}$ is obtained using $\tilde{\boldsymbol{r}}_{c,i}$ and $\tilde{\boldsymbol{\lambda}}_{i}$. 
The origin of the $\{ C_i\}$ coincides with $\tilde{\boldsymbol{r}}_{c,i}$, and by convention, its $z$-axis aligns with $\tilde{\boldsymbol{\lambda}}_{i}$.
In the following, variables expressed in $\{ C_i \}$ are denoted with a left superscript.
We define the $i^{\text{th}}$ environment stiffness matrix with respect to the contact frame by 
\begin{align}
\label{eq:env_stiffness_contact_frame}
^{\{ C_i \}}\boldsymbol{K}_{env,i} = 
\text{diag}(0,0,k_{env,i}),
\end{align}
where $\text{diag}(\cdot)$ represents a diagonal matrix with the given values on its diagonal, and $k_{env,i} \in \mathbb{R}^+$ is the $i^{\text{th}}$ environment stiffness. 
Since only $z$-directional stiffness exists in (\ref{eq:env_stiffness_contact_frame}), any movement of the contact point within the $xy$-plane of the contact frame does not change the force.
This implies that an unconstrained motion can be achieved in the tangential direction of the contact force. 
Subsequently, as shown in Fig.~\ref{fig:spring_contact_feedback}(b), $^{\{ C_i \}}\boldsymbol{r}_{env,i}$ is obtained using  
\begin{align}
\label{eq:env_location_contact_frame}
^{\{ C_i \}}\boldsymbol{r}_{env,i} = ^{\{ C_i \}}\boldsymbol{K}_{env,i} ^{-1} {}^{\{ C_i \}}\tilde{\boldsymbol{\lambda}}_{i}.
\end{align}
Lastly, $\boldsymbol{\theta}_{i}$ is derived from the coordinate transformation from the $i^{\text{th}}$ contact frame to the world frame (see Fig. \ref{fig:spring_contact_feedback}(c)).
Given that $\boldsymbol{R}_i$ represents the rotation matrix for such transformation, $\boldsymbol{\theta}_{i} = \{ \boldsymbol{K}_{env,i}, \boldsymbol{r}_{env,i}\}$ can be obtained as follows:
\begin{align}&
\label{eq:stiff_transform}
\boldsymbol{K}_{env,i} = \boldsymbol{R}_i {}^{\{ C_i \}}\boldsymbol{K}_{env,i} \boldsymbol{R}_i^{T}, \\& \label{eq:env_transform}
\boldsymbol{r}_{env,i} = \tilde{\boldsymbol{r}}_{c,i} + \boldsymbol{R}_i {}^{\{ C_i \}}\boldsymbol{r}_{env,i}.
\end{align}
With this setup, when $\boldsymbol{r}_{c,i} = \tilde{\boldsymbol{r}}_{c,i}$, the contact force computed with (\ref{eq:spring_dyn_1}) and $\boldsymbol{\theta}_{i}$ is the same as $\tilde{\boldsymbol{\lambda}}_{i}$.

The remaining step is to determine the update law for the contact point ($\boldsymbol{r}_{c,i}$).
As depicted in Fig. \ref{fig:spring_contact_feedback}(d), $\boldsymbol{r}_{c,i}$ is fixed on the robot surface, i.e., $\boldsymbol{r}_{c,i}(0) = \boldsymbol{\tilde{r}}_{c,i}$, and it is updated with the robot's forward kinematics.

In summary, when the contact feedback is received, $\boldsymbol{\theta}_{i}$ is computed with (\ref{eq:env_stiffness_contact_frame})-(\ref{eq:env_transform}), $\boldsymbol{r}_{c,i}$ is updated with robot's forward kinematics,
and then the contact force is computed using (\ref{eq:spring_dyn_1}).

\section{Contact Feedback Model Predictive Control}
\label{sec:optimal_control_problem}
Based on the robot system and contact model discussed in Section~\ref{sec:contact_involved_robot_dynamics}, 
this section presents the proposed contact-feedback MPC.

\subsection{Optimal control problem}
\label{subsec:ocp}
The proposed contact-feedback MPC is based on iteratively solving the following discretized optimal control problem (OCP) with the time step $\Delta t $: 
\begin{subequations}
\label{eq:ocp}
\begin{align}
[\boldsymbol{\mathbf{X}}^* , \boldsymbol{\mathbf{U}}^*] = \argmin_{\boldsymbol{\mathbf{X}},\boldsymbol{\mathbf{U}}} \sum_{t=0}^{T-1}L(\boldsymbol{x}^{[t]},\boldsymbol{u}^{[t]})+L_f(\boldsymbol{x}^{[T]}) \label{eq:ocp_objective}
\end{align}
\begin{alignat}{3} \label{eq:ocp_cons1}
&\text{s.t. } \quad && \boldsymbol{x}^{[t+1]} =\boldsymbol{F}(\boldsymbol{x}^{[t]},\boldsymbol{u}^{[t]}, \boldsymbol{\lambda}), \quad &&   \boldsymbol{x}^{[0]} = \tilde{\boldsymbol{x}},\\ \label{eq:ocp_cons2}
&                  \quad && \boldsymbol{\lambda}_{i}^{[t]}=\boldsymbol{g}(\boldsymbol{x}^{[t]}; \boldsymbol{\theta}_i), &&    \\ \label{eq:ocp_cons3}
&                  \quad && \boldsymbol{\theta}_{i} = \boldsymbol{h}(\boldsymbol{\tilde{r}}_{c,i},\boldsymbol{\tilde{\lambda}}_{i}), &&  {\scriptstyle \forall{i} = \{1,2,\ldots,k\}}, \\ \label{eq:ocp_cons4}
&                  \quad && \boldsymbol{x}^{[t]}\in\mathcal{X}, \;    && {\scriptstyle \forall{t} = \{0,1,\ldots,T\}}, \\ \label{eq:ocp_cons5}
&                  \quad && \boldsymbol{u}^{[t]}\in\mathcal{U}, \;  && {\scriptstyle \forall{t} = \{0,1,\ldots,T-1\}},
\end{alignat}
\end{subequations}
where $\boldsymbol{\mathbf{X}}$ and $\boldsymbol{\mathbf{U}}$ denote the state and control input trajectories, e.g., $\boldsymbol{\mathbf{X}} = \{ \boldsymbol{x}^{[0]},\boldsymbol{x}^{[1]},\hdots,\boldsymbol{x}^{[T]} \}$ and $\boldsymbol{\mathbf{U}} = \{ \boldsymbol{u}^{[0]},\boldsymbol{u}^{[1]},\hdots,\boldsymbol{u}^{[T-1]}\}$,
the superscript $*$ indicates that the variable has been optimized,
$L$ and $L_f$ are the discretized running and terminal cost functions, respectively,
and $T$ is the prediction horizon length. 
Here, (\ref{eq:ocp_cons1}) is Euler discretization of (\ref{eq:robot_state_dyn}), 
(\ref{eq:ocp_cons2}) and (\ref{eq:ocp_cons3}) are the spring contact model described in Section~\ref{subsec:spring_dyn}, 
 and the state and control input constraints are considered in (\ref{eq:ocp_cons4}) and (\ref{eq:ocp_cons5}), respectively.
Particularly, in our case, the joint angle, velocity, and control input limits are considered.

\begin{figure}[tb]
\centering
	{\includegraphics[width=0.485\textwidth, trim = {0px 0px 0px 0px}, clip]{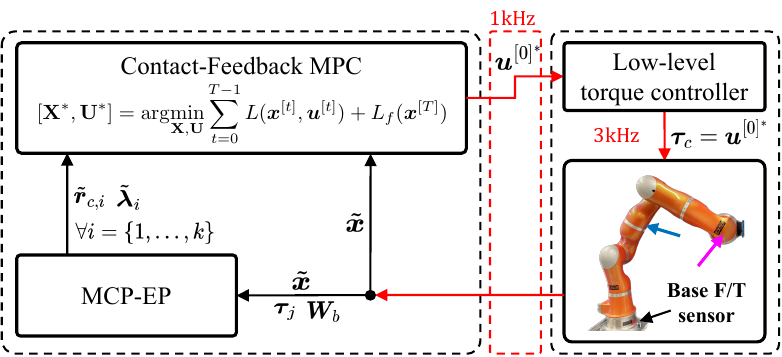}}
	\caption{
 $\boldsymbol{\tau}_j$ and $\boldsymbol{W}_{b}$ are the JTSs and base F/T sensor measurements, respectively, and 
 red arrows represent real-time communication.
 Utilizing the robot states ($\tilde{\boldsymbol{x}}$) and contact information ($\tilde{\boldsymbol{r}}_{c,i}$ and
 $\tilde{\boldsymbol{\lambda}}_{i}$), the contact-feedback MPC (\ref{eq:ocp}) calculates $\boldsymbol{\mathbf{X}}^*$ and 
 $\boldsymbol{\mathbf{U}}^*$. Then, $\boldsymbol{u}^{[0]^*}$ is commanded to the torque-controlled robot.
	}
	\label{fig:control_structure}
	\vspace{-6mm}
\end{figure}

Equation (\ref{eq:ocp}) is solved iteratively, and the first element of the optimal control input trajectory, i.e., $\boldsymbol{u}^{[0]^*}$, is used as a command for a torque-controlled robot. 
A diagram illustrating the proposed contact-feedback MPC framework is shown in Fig. \ref{fig:control_structure}, 
and more details are described in Section~\ref{sec:experiments}.

\subsection{Numerical solution of OCP}
\label{subsec:numerical_sol_ocp}
 
As highlighted in Section~\ref{sec:introduction}, we use the DDP algorithm for (\ref{eq:ocp}) to achieve a real-time MPC.
Specifically, we implemented the method of Box-FDDP \cite{mastalli2022feasibility} to incorporate the spring contact model described in Section~\ref{subsec:spring_dyn}.
This algorithm allows us to impose box-constraints on the control input, e.g., $\boldsymbol{u}_{min} \leq \boldsymbol{u}^{[t]} \leq \boldsymbol{u}_{max}$.
As a result, (\ref{eq:ocp_cons5}) can be treated as hard constraints, whereas (\ref{eq:ocp_cons4}) cannot. To address this limitation, we adopted a penalization approach for (\ref{eq:ocp_cons4}). 
For example, when elements in $\boldsymbol{x}^{[t]}$ exceed the state constraints ($\boldsymbol{x}_{min}, \boldsymbol{x}_{max}$), a steeply increasing quadratic cost is added.
This prevents exceeding the joint and velocity limits during operation.

Additionally, since the DDP algorithm is based on the gradient method, partial derivatives of the cost functions and the contact-involved robot dynamics ($\boldsymbol{f}$) are required.
For faster computation, we derive these analytically rather than using finite-difference or auto-diff methods.
Since $\boldsymbol{r}_{c,i}$ is fixed on the robot surface, the partial derivatives of the contact model (\ref{eq:spring_dyn_1}) with respect to the state and control input are 
\begin{align}
\label{eq:contact_derivative}
\frac{\partial \boldsymbol{\lambda}_{i}}{\partial \boldsymbol{q}} = -\boldsymbol{K}_{env,i}\boldsymbol{J}_{i}(\boldsymbol{q},\boldsymbol{r}_{c,i}), 
&&\frac{\partial \boldsymbol{\lambda}_{i}}{\partial \dot{\boldsymbol{q}}} = \frac{\partial \boldsymbol{\lambda}_{i}}{\partial \boldsymbol{u}} = \boldsymbol{0}.
\end{align}
Consequently, the partial derivatives of $\boldsymbol{f}$ can be analytically derived by combining (\ref{eq:contact_derivative}) with the partial derivatives of the robot forward dynamics that exclude the contact model \cite{carpentier2018analytical}.

\subsection{Cost function design}
\label{subsec:cost_function_design}
In this subsection, we present our cost functions used in (\ref{eq:ocp}) with which the proposed MPC framework is able to accomplish interaction scenarios.
For brevity, $c_{\bullet}\in \mathbb{R}^{+}$ represents the gain of each cost in the following.
Consider the running and terminal cost functions defined by
\begin{align}
\label{eq:running_cost}
    l(\boldsymbol{x},\boldsymbol{u}) &= l_{m}(\boldsymbol{x}) + l_{c}(\boldsymbol{\lambda_{i}}) +c_{u}\norm{\boldsymbol{u}-\boldsymbol{u}_0}^2, \\
    l_f(\boldsymbol{x}) &= l_{m}(\boldsymbol{x}) + l_{c}(\boldsymbol{\lambda_{i}}),
\end{align}
where $l$ and $l_f$ are continuous time version of $L$ and $L_f$ in (\ref{eq:ocp}),
and $\norm{\boldsymbol{u}-\boldsymbol{u}_0}^2$ is the penalization term for the control input.
Here, $l_{c}$ and $l_{m}$ represent the costs for contact force and motion control, respectively.

\subsubsection{Motion control}
\label{subsubsec:running_terminal_cost}
We use the following cost function to penalize the joint velocity and regulate the end-effector of the robot to the desired position and orientation:
\begin{align}
\begin{split}
\label{eq:motion_cost}
    l_{m}(\boldsymbol{x}) &= 
     c_{v}\norm{\dot{\boldsymbol{q}}}^2 +  c_p\norm{\boldsymbol{p}_{ee}(\boldsymbol{q}) - \boldsymbol{p}_{des}}^2\\&
      + c_r\norm{\boldsymbol{R}_{ee}(\boldsymbol{q}) \ominus \boldsymbol{R}_{des}}^2  ,
\end{split}
\end{align}
where $\boldsymbol{p}_{ee}(\boldsymbol{q})$ and $\boldsymbol{R}_{ee}(\boldsymbol{q})$ represent the position and orientation of the end-effector, respectively.
Similarly, $\boldsymbol{p}_{des}$ and $\boldsymbol{R}_{des}$ represent the desired position and orientation,
and $\ominus$ denotes the difference in $\mathrm{SO}(3)$. In particular we used $\boldsymbol{R}_{1} \ominus \boldsymbol{R}_{2}= \log (\boldsymbol{R}_{1}^T\boldsymbol{R}_{2})$.

\subsubsection{Contact force control}
\label{subsubsec:contact_cost}
The cost for contact is designed to allow for defining different tasks on different links (e.g., contact force regulation on link 7 while force barrier on the other links).
To this end, for notational convenience, the $i^{\text{th}}$ contact link is denoted as $\gamma_i \in \{1,2,\hdots,n\}$.

To regulate the $i^{\text{th}}$ contact force to the desired value ($\boldsymbol{\lambda}_{i,des} \in \mathbb{R}^3$), the following cost function is used:
\begin{align}
\label{eq:contact_vec_regulation}
{\small \text{$l_{c,reg}(\boldsymbol{\lambda}_i,\gamma_{set}) =$}}
    \begin{cases}
         {\small \text{$c_{\lambda}\norm{\boldsymbol{A}_{d}(\boldsymbol{\lambda}_i - \boldsymbol{\lambda}_{i,des})}^2$}} & {\scriptsize \text{if $\gamma_{i}\in\gamma_{set}$}}\\
         {\small \text{$0$}} & {\scriptsize \text{otherwise}}
    \end{cases}
    \small{\text{,}}
\end{align}
where $\gamma_{set}$ is the set of link indices used to indicate that this cost is activated,
and $\boldsymbol{A}_d \in \mathbb{R}^{3 \times 3}$ determines the direction of the force control. 
For example, only the $x$-directional force is controlled with $\boldsymbol{A}_d = \text{diag}(1,0,0)$.

For safe handling of unexpected contacts, we limit the allowed contact force.
To this end, the following force barrier with quadratic function is considered:
\begin{align}
\label{eq:contact_norm_barrier}
    &{\small \text{$l_{c,bar}(\boldsymbol{\lambda}_i,\gamma_{set})$}}  \nonumber \\
    &{\small \text{$=$}}
    \begin{cases}
        {\small \text{$c_{\lambda}(\norm{\boldsymbol{\lambda}_i}-\lambda_{i,max})^2$}} &{\scriptsize \text{if $\norm{\boldsymbol{\lambda}_{i}} > b\lambda_{i,max}$ and $\gamma_{i}\in\gamma_{set}$}}\\
        {\small \text{$0$}} &{\scriptsize \text{otherwise}} 
    \end{cases}
    \small{\text{,}}
\end{align}
where $b\in(0,1)$ is the scale factor to smooth the constraint.
As a result, the total cost for the contact $l_{c}(\boldsymbol{x})$ is set as a combination of (\ref{eq:contact_vec_regulation}) and (\ref{eq:contact_norm_barrier}) depending on the applications. 

\section{Experiments}
\label{sec:experiments}
The proposed multi contact-feedback MPC is validated in experiments on the DLR-KUKA LWR IV+ torque-controlled robot.
The low-level torque controller runs at a rate of $3\mathrm{kHz}$.
We use all seven joints of the robot, i.e., $n_x=14$ and $n_u=7$.
To employ the MCP-EP algorithm that estimates contact locations and forces, an ATI Omega F/T sensor is mounted at the base of the robot.
We implemented the proposed contact-feedback MPC in C++, and it operates at approximately $6.8\mathrm{kHz}$, $1.9\mathrm{kHz}$, and $1.8\mathrm{kHz}$ when there are 0, 1, and 2 contacts, respectively.
Since the contact-feedback MPC, MCP-EP, and low-level torque controller operate at different rates, 
each program is run with the latest available data.
Additionally, the update rate of $\boldsymbol{u}^{[0]^*}$ to the low-level torque controller and that of the sensor measurements ($\tilde{\boldsymbol{x}}$, JTSs, base F/T sensor) are consistently set as $1\mathrm{kHz}$ using the real-time setting (see Fig.~\ref{fig:control_structure}).

\subsection{OCP parameter selection}
\label{subsec:ocp_parameter_design}

\subsubsection{Prediction horizon length}
\label{subsubsec:prediction_horizon_length}
In the following experiments, we use a relatively short planning horizon $125\mathrm{ms}$ with $T=5$ and $\Delta t=25\mathrm{ms}$.
It is well known that a longer planning horizon would improve control performance unless it significantly reduces the control frequency.
Despite the fact that, in scenarios involving unexpected contacts, a long planning horizon would not be advantageous.
This is because, as soon as contact occurs, the OCP (\ref{eq:ocp}) should be modified to include $l_c$ in the cost function \eqref{eq:ocp_objective} and to include the spring contact model in \eqref{eq:ocp_cons1}.
This change requires solving an entirely new OCP, distinct from the prior iteration.
This can be problematic for the DDP algorithm, which relies on using the previous step's output as an initial value to enable warm-start.
Specifically, with a long planning horizon, the state trajectory might deviate from the initial state.
This can result in extremely large contact forces in the prediction horizon, leading to numerical issues.

\subsubsection{Environment stiffness}
\label{subsubsec:env_stiffness}
When $k_{env,i}$ has a small value, the robot needs to move significantly within the prediction horizon to control the force.
This can lead to large motion for the actual robot, potentially causing vibrations in practice.
In contrast, when $k_{env,i}$ is too large, each MPC step causes small motion for the actual robot, which may result in too slow convergence of the force.
Finding an appropriate balance between these two is crucial. 
In our implementation, we used $k_{env,i}=3500\mathrm{N/m}$ for all experiments.

Notice again that we have no pre-knowledge about the contacts; instead, the proposed MPC handles the unexpected contacts in real-time.
In the following, two experimental scenarios are presented.
All the experimental videos, including two additional experiments, can be found in the attached file (also see \cite{youtube}).

\begin{figure}[tb!]
\centering
	{\includegraphics[width=0.235\textwidth]{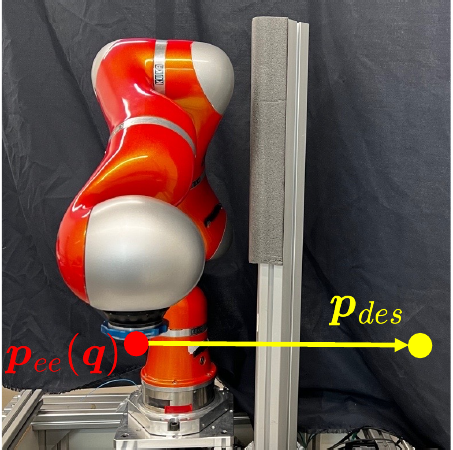}}
	{\includegraphics[width=0.235\textwidth]{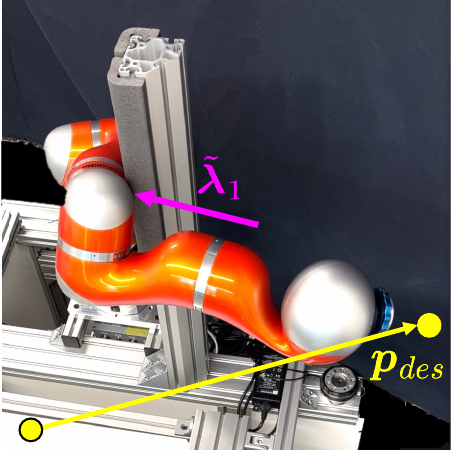}}
	\caption{ 
    Scenario $\#1$:
    The current end-effector position is represented as a red dot, and the desired end-effector position trajectory is given by linear interpolation, as indicated by a yellow arrow.
    While tracking the trajectory, the robot's body makes contact with the obstacle.
	}
	\label{fig:scenario_1}
	\vspace{-6mm}
\end{figure}

\subsection{Comparative study: with and without contact feedback}
\label{subsec:scenario_1}
In scenario $\#1$, we demonstrate the effectiveness of the contact feedback.
To this end, for comparison purposes, we implemented the MPC without contact feedback, in which the OCP (\ref{eq:ocp}) does not include $\boldsymbol{\lambda}$ and $l_c$. For both cases (i.e., with and without contact feedback), the cost for motion control consists of the end-effector position regulation cost and the joint velocity penalization term; i.e., $c_p >0$, $c_v >0$ and $c_r =0$ in \eqref{eq:motion_cost}.

The experimental setup is illustrated in Fig. \ref{fig:scenario_1}.
The robot tracks a desired end-effector position trajectory ($\boldsymbol{p}_{des}$) in the environment where an unknown obstacle is located close to the robot.
$\boldsymbol{p}_{des}$ is set in such a way that the robot cannot reach due to the obstacle. During the trajectory tracking, a contact occurs on the 3rd link at $5.2\mathrm{s}$.

\begin{figure}[tb!]
\centering
    {\includegraphics[width=0.475\textwidth, trim = {6px 0px 13.5px 0px}, clip]{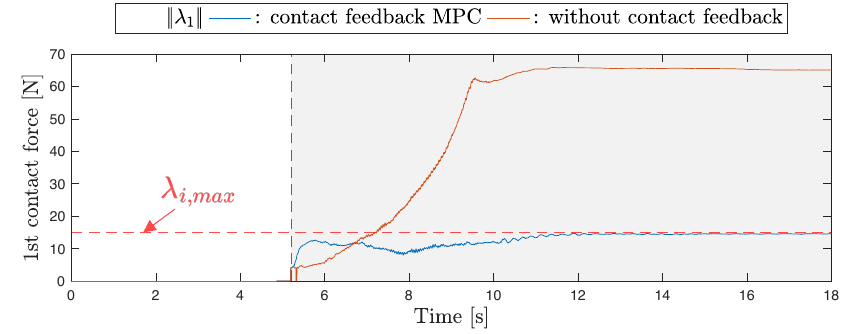}}
    {\includegraphics[width=0.475\textwidth, trim = {6px 2px 13.5px 0px}, clip]{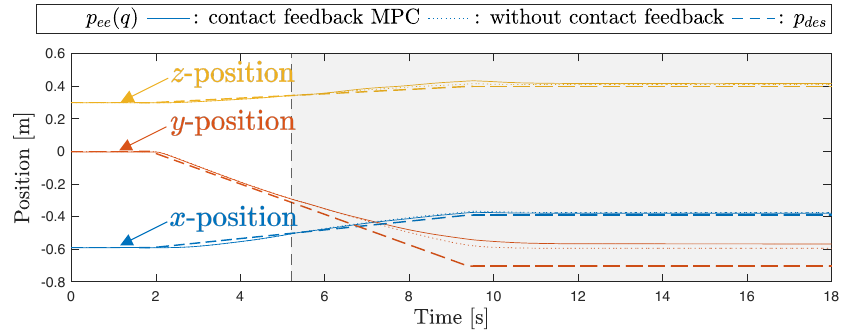}}
	\caption{ 
    Scenario $\#1$: 
    The region is shaded when the contact exists.
    \textbf{Top}: The magnitude of the contact force ($\norm{\tilde{\boldsymbol{\lambda}}_{1}}$).
    \textbf{Bottom}: The actual and desired end-effector trajectories ($\boldsymbol{p}_{ee}$ and $\boldsymbol{p}_{des}$).
	}
	\label{fig:scenario_1_result}
	\vspace{-6mm}
\end{figure}

Fig. \ref{fig:scenario_1_result} shows the contact force ($\norm{\tilde{\boldsymbol{\lambda}}_1}$), the actual end-effector position ($\boldsymbol{p}_{ee}(\boldsymbol{q})) $, and the desired one ($ \boldsymbol{p}_{des}$).
As can be seen in the figure, when there is no contact feedback, the magnitude of the contact force increases as the end-effector position error increases. The maximum contact force was $65.87\mathrm{N}$.

To maintain the interaction force small, the proposed method was implemented using (\ref{eq:contact_norm_barrier}):
\begin{align}
\label{eq:cost_scenario_2}
    l_{c}(\boldsymbol{x}) = \sum_{i=1}^{k}l_{c,bar}(\boldsymbol{\lambda}_i,\gamma_{set}),
\end{align}
where $\gamma_{set} = \{ 1,2,\hdots,7\}$, 
and the upper limit of the contact force ($\lambda_{i,max}$) is set to $15\mathrm{N}$. With the contact feedback, the contact force was maintained below this limit. Over the entire time span, the maximum contact force was $14.74\mathrm{N}$. 

\subsection{Hybrid motion-force control under multi-contact}
\label{subsec:scenario_2_result}

\begin{figure}[tb!]
\centering
        \subfloat[] 
	{\includegraphics[width=0.243\textwidth, trim = {0 0px 0 0px}, clip]{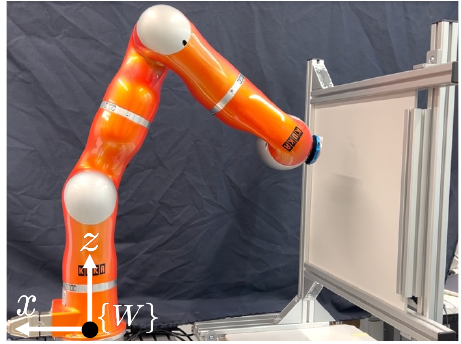}} 
	\subfloat[]
	{\includegraphics[width=0.243\textwidth, trim = {0 0px 0 0px}, clip]{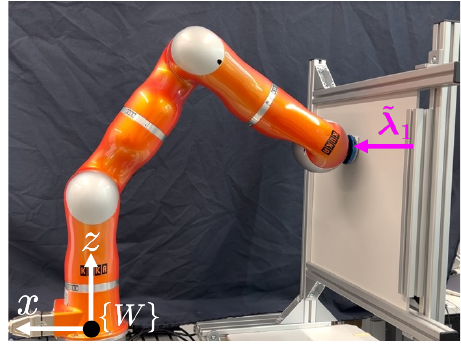}}\\
        \subfloat[]
	{\includegraphics[width=0.243\textwidth, trim = {0 1px 0 0px}, clip]{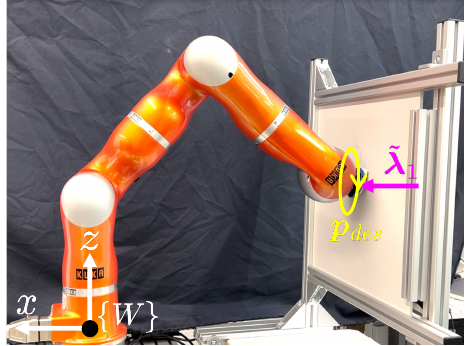}}
	\subfloat[]
	{\includegraphics[width=0.243\textwidth, trim = {0 0px 0px 0px}, clip]{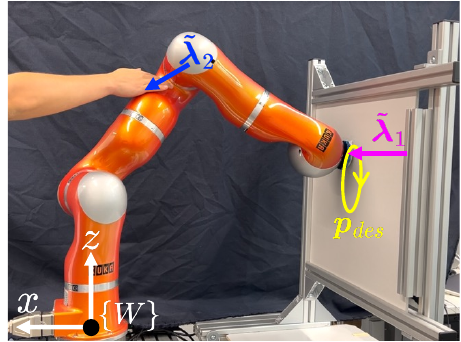}} 
	\caption{ 
    Scenario $\#2$:
    (a) The initial configuration of the robot, not in contact.
    (b) The robot moves in the $x$-direction and makes contact with the environment.
    (c) Hybrid motion-force control is performed at the end-effector.
    (d) While performing hybrid motion-force control, an additional contact is applied by a human. 
	}
	\label{fig:scenario_2}
	\vspace{-6mm}
\end{figure}

In scenario $\#2$, the proposed MPC demonstrates hybrid motion-force control, while handling an unknown interaction.
The experimental scenario is sequentially described in Fig. \ref{fig:scenario_2}.
The robot makes contact with the environment by moving the end-effector along the $x$-direction (see Fig. \ref{fig:scenario_2}(a),(b)).
As shown in Fig. \ref{fig:scenario_2}(c), hybrid motion-force control is performed at the end-effector: the contact force in the $x$-direction is regulated to $20\mathrm{N}$ while tracking a circular trajectory in $yz$-plane.
In addition to the hybrid motion-force control at the end-effector, a human operator applies force on the robot, as shown in Fig.~\ref{fig:scenario_2}(d).

In this scenario, the proposed method is expected to handle unknown contact safely, while maintaining $20\mathrm{N}$ at the end-effector. To this end, the cost for the contact is defined as follows:
\begin{align}
\label{eq:cost_scenario_1}
    l_{c}(\boldsymbol{x}) = \sum_{i=1}^{k}l_{c,bar}(\boldsymbol{\lambda}_i,\gamma_{set,1}) + l_{c,reg}(\boldsymbol{\lambda}_i,\gamma_{set,2}),
\end{align}
where $\gamma_{set,1} = \{ 1,2,\hdots,6\}$, $\lambda_{i,max}=15\mathrm{N}$, $\gamma_{set,2} = \{7\}$, $\boldsymbol{A}_d = \text{diag}(1,0,0)$, and the first element of $\boldsymbol{\lambda}_{i,des}$ is $20\mathrm{N}$. 

To track the desired trajectory, a high gain was set for $c_p$, while a slightly lower value was allocated to the orientation part $c_r$. The joint velocity penalization term was also used; i.e., $c_v>0$.
Note that with the contact model presented in Section~\ref{subsec:spring_dyn}, robot motion can be achieved in the unconstrained tangential direction of the contact force, allowing for hybrid motion-force control.

\begin{figure}[tb!]
\centering
	{\includegraphics[width=0.475\textwidth, trim = {6px 0px 13.5px 1px}, clip]{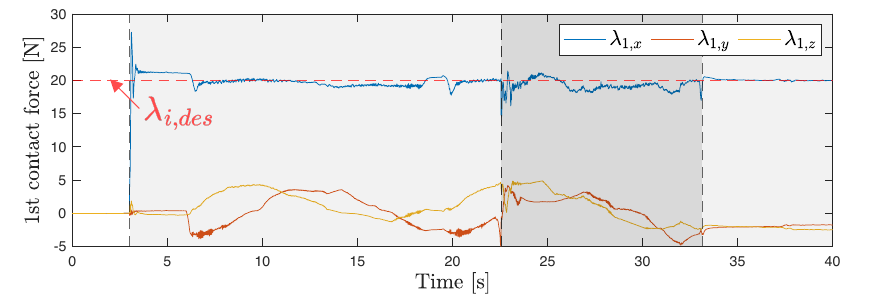}}
        {\includegraphics[width=0.475\textwidth, trim = {5.5px 0px 14.5px 1px}, clip]{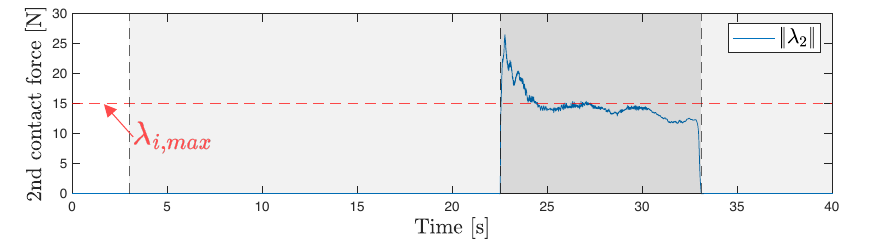}}
        {\includegraphics[width=0.475\textwidth, trim = {5.5px 1.5px 14.5px 1px}, clip]{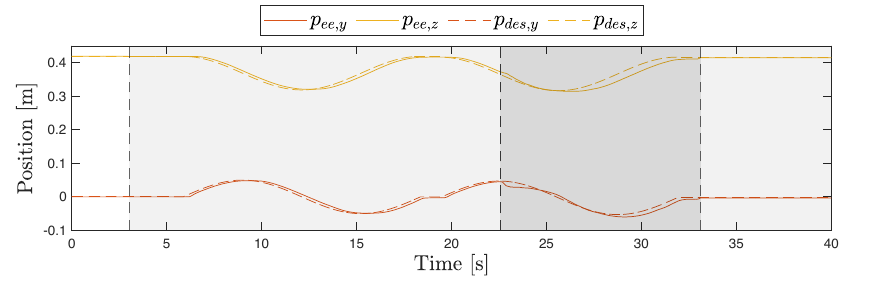}}
	\caption{ Scenario $\#2$: 
        When there are two contacts, the region is shaded more intensely.
        \textbf{Top}: The contact force at the end-effector ($\tilde{\boldsymbol{\lambda}}_{1}$).
        \textbf{Middle}: The magnitude of the contact force at the robot's body ($\norm{\tilde{\boldsymbol{\lambda}}_{2}}$).
        \textbf{Bottom}: The actual and desired end-effector trajectories ($\boldsymbol{p}_{ee}$ and $\boldsymbol{p}_{des}$).
	}
	\label{fig:scenario_2_result}
	\vspace{-6mm}
\end{figure}

Similar to the previous, Fig. \ref{fig:scenario_2_result} shows the first contact force ($\tilde{\boldsymbol{\lambda}}_1$), the second contact force ($\norm{\tilde{\boldsymbol{\lambda}}_2}$), the end-effector's actual position ($\boldsymbol{p}_{ee}(\boldsymbol{q}) $), and the desired position ($ \boldsymbol{p}_{des}$).
At $3\mathrm{s}$, the end-effector makes contact with the environment.
From $5.9 \mathrm{s} $ to $18.4 \mathrm{s} $, only the hybrid motion-force control at the end-effector is performed without human interaction.
In this case, the RMSE of the end-effector position was $0.86\mathrm{cm}$, and that of the contact force was $0.58\mathrm{N}$.

On the other hand, while tracking the second circular trajectory (from $19.4 \mathrm{s} $ to $31.9 \mathrm{s}$), additional contact force ($\tilde{\boldsymbol{\lambda}}_{2}$) is applied to the 3rd link by a human.
As shown in Fig. \ref{fig:scenario_2_result}, the robot immediately reacts to maintain the contact force below $15\mathrm{N}$. 
At the same time, the motion-force control at the end-effector performs reasonably. 
Specifically, at the end-effector, the RMSE for the position was $1.14\mathrm{cm}$ and for the contact force was $0.89\mathrm{N}$.


Finally, from $33.5\mathrm{s}$ to the end, the force regulation control is performed with a constant desired pose.
The RMSE values for the end-effector position and contact force were $0.20\mathrm{cm}$ and $0.10\mathrm{N}$, respectively.

\section{Conclusion and Future Work}
\label{sec:conclusion}

In this work, we propose a contact-feedback MPC that accomplishes interactive robotic tasks in which multiple contacts may occur at unknown locations.
To accomplish such scenarios, we adopt a spring contact model to predict the contact-involved robot motion over the prediction horizon of the MPC.
Here, the multi-contact information, that initializes the spring contact model, is provided by our previous work called MCP-EP.
To the best of our knowledge, this paper presented the MPC that explicitly handles unknown multiple contacts for the first time.
Furthermore, since the proposed method achieves fast enough runtime for a 7-DOF robot (update rate was $1.9\mathrm{kHz}$ for the single-contact, and $1.8\mathrm{kHz}$ for the dual-contact), the robot is able to handle unexpected contacts swiftly.
The effectiveness of the proposed method is validated through the real-world experiments using a 7-DOF DLR-KUKA LWR IV+.

Although our approach has shown promising results, there remains room for improvement.
When the contact feedback is received, the contacts remain throughout the entire prediction horizon of the OCP.
However, since this contact model is derived from the contact feedback, its validity is limited only locally.
For example, when employing a long horizon length, the issue with extremely high contact force can arise as discussed in Section~\ref{subsubsec:prediction_horizon_length}.
Moreover, the spring contact model used in this paper does not account for instances where the contact might vanish.
Considering these issues could allow a longer prediction horizon in the OCP, potentially improving overall controller performance.
Furthermore, to truly address the unexpected contacts, online determination of $k_{env,i}$ is essential, and therefore, this would be our future work as well.

\bibliographystyle{myIEEEtran.bst}
\bibliography{IEEEabrv,[bib]ICRA2024_ddp.bib}

\begin{thebibliography}{10}
\providecommand{\url}[1]{#1}
\csname url@samestyle\endcsname
\providecommand{\newblock}{\relax}
\providecommand{\bibinfo}[2]{#2}
\providecommand{\BIBentrySTDinterwordspacing}{\spaceskip=0pt\relax}
\providecommand{\BIBentryALTinterwordstretchfactor}{4}
\providecommand{\BIBentryALTinterwordspacing}{\spaceskip=\fontdimen2\font plus
\BIBentryALTinterwordstretchfactor\fontdimen3\font minus \fontdimen4\font\relax}
\providecommand{\BIBforeignlanguage}[2]{{%
\expandafter\ifx\csname l@#1\endcsname\relax
\typeout{** WARNING: IEEEtran.bst: No hyphenation pattern has been}%
\typeout{** loaded for the language `#1'. Using the pattern for}%
\typeout{** the default language instead.}%
\else
\language=\csname l@#1\endcsname
\fi
#2}}
\providecommand{\BIBdecl}{\relax}
\BIBdecl

\bibitem{iskandar2023hybrid}
M.~Iskandar, C.~Ott, A.~Albu-Sch{\"a}ffer, B.~Siciliano, and A.~Dietrich, ``Hybrid force-impedance control for fast end-effector motions,'' \emph{IEEE Robotics and Automation Letters}, vol.~8, no.~7, pp. 3931--3938, 2023.

\bibitem{ott2010unified}
C.~Ott, R.~Mukherjee, and Y.~Nakamura, ``Unified impedance and admittance control,'' in \emph{IEEE International Conference on Robotics and Automation (ICRA)}, 2010, pp. 554--561.

\bibitem{kim2019model}
M.~J. Kim, F.~Beck, C.~Ott, and A.~Albu-Sch{\"a}ffer, ``Model-free friction observers for flexible joint robots with torque measurements,'' \emph{IEEE Transactions on Robotics}, vol.~35, no.~6, pp. 1508--1515, 2019.

\bibitem{9812268}
J.~Jeong, H.~Mishra, C.~Ott, and M.~J. Kim, ``A memory-based {SO(3)} parameterization: Theory and application to {6D} impedance control with radially unbounded potential function,'' in \emph{IEEE International Conference on Robotics and Automation (ICRA)}, 2022, pp. 8338--8344.

\bibitem{9643422}
M.~J. Kim, A.~Werner, F.~Loeffl, and C.~Ott, ``Passive impedance control of robots with viscoelastic joints via inner-loop torque control,'' \emph{IEEE Transactions on Robotics}, vol.~38, no.~1, pp. 584--598, 2022.

\bibitem{7759143}
M.~J. Kim, W.~Lee, C.~Ott, and W.~K. Chung, ``A passivity-based admittance control design using feedback interconnections,'' in \emph{IEEE/RSJ International Conference on Intelligent Robots and Systems (IROS)}, 2016, pp. 801--807.

\bibitem{keemink2018admittance}
A.~Q. Keemink, H.~van~der Kooij, and A.~H. Stienen, ``Admittance control for physical human--robot interaction,'' \emph{The International Journal of Robotics Research}, vol.~37, no.~11, pp. 1421--1444, 2018.

\bibitem{8695069}
M.~J. Kim, W.~Lee, J.~Y. Choi, G.~Chung, K.-L. Han, I.~S. Choi, C.~Ott, and W.~K. Chung, ``A passivity-based nonlinear admittance control with application to powered upper-limb control under unknown environmental interactions,'' \emph{IEEE/ASME Transactions on Mechatronics}, vol.~24, no.~4, pp. 1473--1484, 2019.

\bibitem{tassa2012synthesis}
Y.~Tassa, T.~Erez, and E.~Todorov, ``Synthesis and stabilization of complex behaviors through online trajectory optimization,'' in \emph{IEEE/RSJ International Conference on Intelligent Robots and Systems (IROS)}, 2012, pp. 4906--4913.

\bibitem{tassa2014control}
Y.~Tassa, N.~Mansard, and E.~Todorov, ``Control-limited differential dynamic programming,'' in \emph{IEEE International Conference on Robotics and Automation (ICRA)}, 2014, pp. 1168--1175.

\bibitem{mastalli2022feasibility}
C.~Mastalli, W.~Merkt, J.~Marti-Saumell, H.~Ferrolho, J.~Sol{\`a}, N.~Mansard, and S.~Vijayakumar, ``A feasibility-driven approach to control-limited {DDP},'' \emph{Autonomous Robots}, vol.~46, no.~8, pp. 985--1005, 2022.

\bibitem{aoyama2021constrained}
Y.~Aoyama, G.~Boutselis, A.~Patel, and E.~A. Theodorou, ``Constrained differential dynamic programming revisited,'' in \emph{IEEE International Conference on Robotics and Automation (ICRA)}, 2021, pp. 9738--9744.

\bibitem{kleff2022introducing}
S.~Kleff, E.~Dantec, G.~Saurel, N.~Mansard, and L.~Righetti, ``Introducing force feedback in model predictive control,'' in \emph{IEEE/RSJ International Conference on Intelligent Robots and Systems (IROS)}, 2022, pp. 13\,379--13\,385.

\bibitem{mastalli2020crocoddyl}
C.~Mastalli, R.~Budhiraja, W.~Merkt, G.~Saurel, B.~Hammoud, M.~Naveau, J.~Carpentier, L.~Righetti, S.~Vijayakumar, and N.~Mansard, ``Crocoddyl: An efficient and versatile framework for multi-contact optimal control,'' in \emph{IEEE International Conference on Robotics and Automation (ICRA)}, 2020, pp. 2536--2542.

\bibitem{kazim2018combined}
K.~J. Kazim, J.~Bethge, J.~Matschek, and R.~Findeisen, ``Combined predictive path following and admittance control,'' in \emph{Annual American Control Conference (ACC)}, 2018, pp. 3153--3158.

\bibitem{wahrburg2016mpc}
A.~Wahrburg and K.~Listmann, ``{MPC}-based admittance control for robotic manipulators,'' in \emph{IEEE 55th Conference on Decision and Control (CDC)}, 2016, pp. 7548--7554.

\bibitem{bednarczyk2020model}
M.~Bednarczyk, H.~Omran, and B.~Bayle, ``Model predictive impedance control,'' in \emph{IEEE international conference on robotics and automation (ICRA)}, 2020, pp. 4702--4708.

\bibitem{gold2020model}
T.~Gold, A.~V{\"o}lz, and K.~Graichen, ``Model predictive interaction control for industrial robots,'' \emph{IFAC-PapersOnLine}, vol.~53, no.~2, pp. 9891--9898, 2020.

\bibitem{gold2021model}
------, ``Model predictive position and force trajectory tracking control for robot-environment interaction,'' in \emph{IEEE/RSJ International Conference on Intelligent Robots and Systems (IROS)}, 2020, pp. 7397--7402.

\bibitem{gold2022model}
------, ``Model predictive interaction control for robotic manipulation tasks,'' \emph{IEEE Transactions on Robotics}, vol.~39, no.~1, pp. 76--89, 2022.

\bibitem{matschek2017force}
J.~Matschek, J.~Bethge, P.~Zometa, and R.~Findeisen, ``Force feedback and path following using predictive control: Concept and application to a lightweight robot,'' \emph{IFAC-PapersOnLine}, vol.~50, no.~1, pp. 9827--9832, 2017.

\bibitem{matschek2020direct}
J.~Matschek, R.~Jordanowa, and R.~Findeisen, ``Direct force feedback using {Gaussian} process based model predictive control,'' in \emph{IEEE Conference on Control Technology and Applications (CCTA)}, 2020, pp. 8--13.

\bibitem{matschek2023safe}
J.~Matschek, J.~Bethge, and R.~Findeisen, ``Safe machine-learning-supported model predictive force and motion control in robotics,'' \emph{IEEE Transactions on Control Systems Technology}, vol.~31, no.~6, pp. 2380--2392, 2023.

\bibitem{iskandar2021collision}
M.~Iskandar, O.~Eiberger, A.~Albu-Sch{\"a}ffer, A.~De~Luca, and A.~Dietrich, ``Collision detection, identification, and localization on the dlr sara robot with sensing redundancy,'' in \emph{IEEE International Conference on Robotics and Automation (ICRA)}, 2021, pp. 3111--3117.

\bibitem{manuelli2016localizing}
L.~Manuelli and R.~Tedrake, ``Localizing external contact using proprioceptive sensors: The contact particle filter,'' in \emph{IEEE/RSJ International Conference on Intelligent Robots and Systems (IROS)}, 2016, pp. 5062--5069.

\bibitem{10161173}
S.~W. Han and M.~J. Kim, ``Proprioceptive sensor-based simultaneous multi-contact point localization and force identification for robotic arms,'' in \emph{IEEE International Conference on Robotics and Automation (ICRA)}, 2023, pp. 12\,099--12\,105.

\bibitem{pang2022easing}
T.~Pang and R.~Tedrake, ``Easing reliance on collision-free planning with contact-aware control,'' in \emph{IEEE International Conference on Robotics and Automation (ICRA)}, 2022, pp. 8375--8381.

\bibitem{zube2016model}
A.~Zube, J.~Hofmann, and C.~Frese, ``Model predictive contact control for human-robot interaction,'' in \emph{Proceedings of ISR: 47st International Symposium on Robotics}.\hskip 1em plus 0.5em minus 0.4em\relax VDE, 2016, pp. 1--7.

\bibitem{8695719}
J.~M. Gandarias, A.~J. García-Cerezo, and J.~M. Gómez-de Gabriel, ``{CNN}-based methods for object recognition with high-resolution tactile sensors,'' \emph{IEEE Sensors Journal}, vol.~19, no.~16, pp. 6872--6882, 2019.

\bibitem{8932392}
J.~Liang, J.~Wu, H.~Huang, W.~Xu, B.~Li, and F.~Xi, ``Soft sensitive skin for safety control of a nursing robot using proximity and tactile sensors,'' \emph{IEEE Sensors Journal}, vol.~20, no.~7, pp. 3822--3830, 2020.

\bibitem{jain2013reaching}
A.~Jain, M.~D. Killpack, A.~Edsinger, and C.~C. Kemp, ``Reaching in clutter with whole-arm tactile sensing,'' \emph{The International Journal of Robotics Research}, vol.~32, no.~4, pp. 458--482, 2013.

\bibitem{killpack2016model}
M.~D. Killpack, A.~Kapusta, and C.~C. Kemp, ``Model predictive control for fast reaching in clutter,'' \emph{Autonomous Robots}, vol.~40, pp. 537--560, 2016.

\bibitem{carpentier2018analytical}
J.~Carpentier and N.~Mansard, ``Analytical derivatives of rigid body dynamics algorithms,'' in \emph{Robotics: Science and systems (RSS)}, 2018.

\bibitem{youtube}
\BIBentryALTinterwordspacing
{KAIST IRSL}, ``{Multi-Contact Feedback MPC for interactive robotic tasks (ICRA2024)},'' (Feb. 27, 2024). Accessed: Feb. 27, 2024. [Online Video]. Available: \url{https://youtu.be/pZbkHaxT1ao}
\BIBentrySTDinterwordspacing

\end{thebibliography}

\end{document}